\documentclass[10pt,onecolumn,letterpaper]{IEEEtran}
\usepackage{spconf,amsmath,epsfig}
\usepackage{amsmath}
\usepackage{subfigure}
\usepackage{epstopdf}
\usepackage{graphicx}
\usepackage{amsfonts}
\usepackage{multirow}
\usepackage{wrapfig}
\usepackage{subfloat}
\usepackage{lipsum}
\usepackage[lined,boxed,commentsnumbered]{algorithm2e}
\usepackage{color}
\usepackage{subfig}
\usepackage{makeidx}
\usepackage{moreverb}

\usepackage{array}
\usepackage{bbding}
  \DeclareGraphicsExtensions{.eps,.jpg,.png,.tif}

\begin{document}
\title{Fusion of Image Segmentation Algorithms using Consensus Clustering} %

\name{ Mete~Ozay $\dag \ddag$, Fatos~T.~Yarman~Vural ${\ddag}$, Sanjeev~R.~Kulkarni ${\dag}$, and~H.~Vincent~Poor ${\dag}$}

 \address{${\dag}$ Department of Electrical Engineering, Princeton University, NJ 08544 USA \\ ${\ddag}$ Department of Computer Engineering, Middle East Technical University, Ankara, Turkey}%

\maketitle

\begin{abstract}
A new segmentation fusion method is proposed that ensembles the output of several segmentation algorithms applied on a remotely sensed image. 
The candidate segmentation sets are processed to achieve a consensus segmentation using a stochastic optimization algorithm based on the  Filtered Stochastic BOEM (Best One Element Move) method. For this purpose, Filtered Stochastic BOEM is reformulated as a segmentation fusion problem by designing a new distance learning approach. The proposed algorithm also embeds the computation of the optimum number of clusters into the segmentation fusion problem. 
\end{abstract}

%
\section{Introduction}

In hyper-spectral remote sensing problems, it is difficult to find an optimal segmentation algorithm that covers all the spectral bands. Some objects are recognized on specific spectral bands, whereas other objects may require the processing of different bands together. For example, the algorithms with a set of selected parameters may successfully detect objects such as water and shadow in the near-infrared (NIR) band, but may fail to detect objects which provide color or textural information, such as farms and buildings. Therefore, one may need to employ more than one segmentation output obtained from multiple spectral bands to extract various types of objects. Additionally, depending on the object types, one may need to employ more than one set of features in the segmentation algorithms. 
 
In this study, we introduce a new approach for the segmentation fusion problem based on a consensus clustering algorithm, called Stochastic Filtered Best One Element Move (Filtered Stochastic BOEM) \cite{Zheng}. The proposed method can also be employed to find the optimal set of parameters for a segmentation algorithm for a dataset. We first employ different segmentation algorithms or a single segmentation algorithm with a set of different parameters to a remote sensing image and obtain a set of candidate outputs. Then, we design a fusion strategy by adapting the Filtered Stochastic BOEM method. There are two major contributions of the proposed segmentation fusion method. The first is to formalize the Filtered Stochastic BOEM method as a segmentation fusion problem, where we design a new distance learning method. The second contribution is to embed the computation of the optimal cluster number into the Filtered Stochastic BOEM method. In the suggested framework, we assume that some of the segments in the candidate segmentation set are expected to represent acquired target objects.

Three well-known segmentation algorithms, k-means, Graph Cuts \cite{gc1,gc2,gc3} and Mean Shift \cite{fuku,ms} are used as the base segmentation algorithms in order to segment benchmark hyperspectral image datasets. In the next section, we introduce our segmentation fusion method. We examine the suggested method
with various experiments in Section 3. Section 4 concludes the paper.
\section{Filtered Stochastic BOEM Formulated for the Fusion of Segmentation Algorithms}

Filtered Stochastic BOEM \cite{Zheng} is a consensus clustering algorithm which approximates a solution to the Median Partition Problem \cite{BOEM} by integrating BOEM \cite{BOEM} and Stochastic Gradient Descent (SGD) \cite{sg}. 

In the proposed segmentation fusion method, we first feed an image $I$ to $D$ segmentation algorithms $SA_j$, $j=1,...,D$. Each segmentation algorithm is employed on $I$ to obtain a set of segmentation outputs $S_j= \{ s_{i} \} ^{k_j} _{i=1} $ where $s_i \in A ^N$ is a segmentation output, $A$ is the set of segment labels with $N$ pixels with $|A|=C$ different segment labels, and a distance function $d(\cdotp, \cdotp)$. 

An initial segmentation $s$ is selected from the segmentation set $S=\bigcup\limits ^D _{j=1} S_j$ consisting of $K=\sum ^D _{j=1} k_j$ segmentations using algorithms which employ search heuristics, such as Best of K \textit{(BOK)} \cite{BOEM}. Then, a \textit{consensus} segmentation $\hat { s }$ is computed by solving the following optimization problem:
\[
\hat { s } = \underset{ s }{\mathrm{argmin}} \sum\limits ^{K} _{i=1} d( s _i , s ) \; . 
\]

Given two segmentations $s_i$ and $s_j$, the distance function is defined as the \textit{symmetric distance function (SDD)} given by $d( s_i , s_j)=N_{01} + N_{10}$, where $N_{01}$ is the number of pairs co-segmented in $s_i$ but not in $s_j$, and $N_{10}$ is the number of pairs co-segmented in $s_j$ but not in $s_i$. In order to compare segmentations with a different number of pixels $N$ and segmentations $K$, we use a normalized form of $SDD$ which is called \textit{Average Sum of Distances (Average SOD)}
\begin{equation}
\frac{2 \sum ^K _{i=1} d( s_i , s)}{KN(N-1)} \; \; \; .\label{eq:sod}
\end{equation}

At each iteration of the optimization algorithm, a new segmentation is computed. Specifically, a segmentation $s$ is randomly selected from the
segmentation set. Then, the best one element move of the current segmentation $s$ is computed with respect to the objective of the optimization and applied to the current segmentation to generate a new segmentation. If there is no improvement on the best move, the current segmentation is returned by the algorithm.

Similar to the gradient descent method, the best one element move of segmentation $s$ is defined as
\[
\Delta s = \frac{ \partial \sum\limits ^{K} _{i=1} d( s _i , s ) } { \partial s } ,
\]
and can be evaluated by $ \Delta s _t = \frac{ \partial H_t } { \partial s _t } $, where $H_t = \sum\limits ^{K} _{i=1} d( s _i , s_t ) $ is the objective at time $t$. Using the assumption that single element updates do not change the objective function, $H_t$ can be approximated by $H_{t-1}$ with a scale parameter $ \beta \in (0,1) $. Then,
\[
\Delta s _t = \frac{ \partial } { \partial s _t } ( \beta H_{t-1} + d( s_k, s_t) ) \; ,
\]
where $s_k$ is the randomly selected segmentation for updating the current BOEM. If an $N \times C$ matrix $[H]$ is defined such that the $i ^{th}$ row and the $j ^{th}$ column of the matrix, $[H] _{ij}$, is the updated value of $H$ obtained by switching $i ^{th}$ element of $s$ to the $j ^{th}$ segment label, the move can be approximated by
\begin{equation}
\underset{ i,j }{\mathrm{argmin}} \; \beta [H_{t-1} ]_{i,j} + [d( s _k , s _t )]_{i,j} \; , 
\label{eq:app}
\end{equation}
if the $k^{th}$ segmentation is selected for updating $s _t$ at time $t$.

\begin{algorithm}
%
\SetKwFunction{Union}{Union}\SetKwFunction{FindCompress}{FindCompress}
\SetKwInOut{Input}{input}\SetKwInOut{Output}{output}
\Input {Input image $I$, $\{ SA_j \} ^D _{j=1}$, $T$. }
\Output{Output image $O$}
\nl Run $SA_j$ on $I$ to obtain $S_j= \{ s_{i} \} ^{k_j} _{i=1}$, $\forall j=1,...,D.$

\nl At $t=1$, initialize $s$ and $ [H_t]$

\For{$t \leftarrow 2$ \KwTo $T$}{
\nl Randomly select one of the segmentation results $k \in \{ 1,2,...,K \}$

\nl $ [H_t] \leftarrow \beta [H_t] + [d( s_k, s) ]$ 

\nl Find $\Delta s$ by solving $\underset{ i,j }{\mathrm{argmin}} \beta [H_{t} ]_{i,j}$

\nl $ s \leftarrow s + \Delta s$

\nl $ t \leftarrow t+1$ }

\nl $O \leftarrow s$ 

\caption{Segmentation Fusion}
\end{algorithm}

In the proposed Segmentation Fusion Algorithm, we initialize $ [H_t]$ at $t=1$. Until $t$ reaches a given termination time $T$, we update the segmentation $s$. We randomly select a segmentation from a pseudo-random permutation of the numbers $1,2,...,K$ until we traverse all the segmentations in $s_1, s_2,...,s_K$. Then, we generate a new segmentation and repeat this operation until all of the permutations are traversed. We update $[H_t]$ by aggregating $[d( s_k, s) ]$ with the scaled $ \beta [H_t]$. $\beta \in [0,1]$ controls the convergence rate and the performance of the algorithm. If $\beta =0$, the algorithm becomes pure stochastic BOEM and the algorithm is memoryless. If $\beta =1$, the algorithm \textit{forgets} slowly. However, Zheng, Kulkarni and Poor \cite{Zheng} reported that the algorithm may perform worse if $\beta$ is on either end of $[0,1]$. Selection of  the \textit{optimal} $\beta$ values for segmentation fusion is explained in the next section. After $[H_t]$ is updated, we compute $\Delta s$ in order to update $s$. We iterate the algorithm until the termination criterion is achieved.  

\subsection{Distance Learning}

In this section, we propose a method, called distance learning that employs the training data to measure the distance between two segmentations obtained at the output of different segmentation algorithms. The proposed distance learning method is also flexible for measuring the distance between two segmentations with different numbers of segments. 

We first define \textit{Rand Index} ($RI$), which is used to estimate the quality of the segments. Given two segmentations $s_i$ and $s_j$, $RI$ is defined as $ RI(s_i, s_j)=1-\frac{d( s_i, s_j)}{\binom{N}{2}}$, where $d( s_i, s_j)= N_{10}+N_{01}=\binom{N}{2} - (N_{00}+N_{11})$. However, $RI$ is not corrected for chance, for instance, the average distance between two segmentations is not zero and the distance depends on the number of pixels \cite{ARI}. Therefore, we assume that each segmentation $s_i = \{ s_{i,k_i} \} ^{K_i} _{k_i=1} $ consists of different numbers of segments $K_i$. We define $\aleph(i)$ as the number of pixels in the $i^{th}$ segment of $s_i$, and  $\aleph_{ij}$ as the number of pixels in both the $i^{th}$ segment of $s_i$ and the $j^{th}$ segment of $s_j$. In addition, we assume that $s_i$ and $s_j$ are randomly drawn with a fixed number of segments, and a fixed number pixels in each segment according to a generalized hypergeometric distribution \cite{kuncheva}. Then, an adjusted version of $RI$ called \textit{Adjusted Rand Index} ($ARI$) \cite{kuncheva} is defined as

\begin{equation}
ARI(s_i, s_j)=\frac{\sum ^{K_i} _ {k_i=1} \sum ^{K_j} _ {k_j=1} \binom{\aleph_{ij}}{2} -\theta_{ij}}{ \frac{1}{2} ( \theta_i + \theta_j ) - \theta_{ij} } \; ,
\label{eq:ARI}
\end{equation}
where $\theta_i= \sum ^{K_i} _ {k_i=1} \binom{\aleph(i)}{2}$,  $\theta_j= \sum ^{K_j} _ {k_j=1} \binom{\aleph(j)}{2}$ and \newline $\theta_{ij}=\frac{2 \theta_i \theta_j}{N(N-1)}$.

Note that, if we apply our assumptions for equal segmentation sizes $K_i=K_j, \forall i \neq j$ in $ARI$, we obtain \eqref{eq:sod} \cite{BOEM}. Instead, we compute $ARI( s_i, s_j)$ for each different base segmentation algorithm output $S_i$ with different segment numbers $K_i$ and $d( s_i, s_j)$ is computed from $ARI( s_i, s_j)$, such that $d( s_i, s_j) = 1-ARI( s_i, s_j)$ \cite{BOEM}. We call this method Distance Learning for BOEM (DL) in which we learn $d( s_i, s_j)$ by computing $ARI( s_i, s_j)$ using the data. 

An important assumption that is made in the derivation of $ARI$ \cite{Info_theoretic} is that the number of pixels in each segment is the same. However, this assumption may fail in the segmentation of images that contain complex targets, such as airports or harbors. 

In order to relax this assumption, we employ a normalization method for quasi-distance functions, introduced by Luo et al. \cite{quasi} as
\begin{equation}
nd(s_i, s_j)=\frac{d( s_i, s_j) - d_{min}( s_i, s_j)}{d_{max}( s_i, s_j) - d_{min}( s_i, s_j)} \; ,
\label{eq:quasi}
\end{equation}
where $d_{min}( s_i, s_j)$ and $d_{max}( s_i, s_j)$ are the minimal and maximal values of $d( s_i, s_j)$. Luo et al. \cite{quasi} states that the exact computation of $d_{max}( s_i, s_j)$ for any segmentation distribution is not known and they introduce several approximations. In the experiments, we employ \eqref{eq:quasi} as the method called Quasi-distance Learning(QD). For the details of the algorithms to solve \eqref{eq:quasi}, please refer to \cite{quasi}.

An important difference between \eqref{eq:ARI} and \eqref{eq:quasi} is that we consider the minimal and maximal values of the distances between the pairwise segmentations $(s_i, s_j)$ as the normalization factors, in order to compute the distances between $s_i$ and $s_j$, in \eqref{eq:quasi}. On the other hand, \eqref{eq:ARI} considers the expected values of the distances between all of the segmentations in the computations.  

If the training data is available, then $d_{min}( s_i, s_j)$ and $d_{max}( s_i, s_j)$ can be computed using the training data and employed to test data. However, one must assure that the statistical properties of training and test data are equivalent in order to employ the learning methods. We observe that this equivalent requirement may not be satisfied in remote sensing datasets in the experiments because of the variability of the images in the context of space and time.

\subsection{Estimating Number of Clusters and parameters $\beta$ for BOEM}

One of the crucial problems of image segmentation is to estimate the number of clusters that forms different segments, $C$, in the image. This problem is very crucial for the segmentation of remotely sensed images even if the images are labeled using expert knowledge. 

In order to estimate $C$ in the base segmentation algorithms, several clustering validity indices can be employed \cite{k1}. In this section, we introduce a new method to estimate $C$ for segmentation fusion. For this purpose, we consider a segmentation index (SI) for BOEM as $SI(c)= \sum _{i<j} ARI(s_i, s_j)$, where $\{ s_i \} ^K _{i=1}$ is the set of $K$ segmentations where each segmentation $s_i$ contains segments with $c$ different labels \cite{kboem}. Then, we solve the following optimization problem,
\begin{equation}
\hat{C}= \underset{ c=2,...,C_{max} }{\mathrm{argmin}} \; SI(c) \; ,
\label{eq:findc}
\end{equation}
where $C_{max}$ is the maximum value of $c$ provided by the user. Vinh and Epps \cite{kboem} compared Normalized Mutual Information and $ARI$ for the estimation of segment number on several datasets. Since both of the algorithms agree on the segment number in various experiments, we employ $ARI$ in our experiments for estimating $c$.

A similar approach is employed to estimate the parameter $\beta$. Given a set of $\beta$ values $\Xi= \{ \beta_b \} ^B _{b=1}$, we introduce a beta index ($BI$) as $
BI( \beta_b)= \sum ^K _{i=1} ARI(s_i, O (\beta_b) )$, where $O (\beta_b)$ is the output segmentation of the \textbf{Segmentation Fusion Algorithm} implemented using $\beta_b$. Then the optimal $\hat{\beta}$ is computed by solving the following optimization algorithm:
\begin{equation}
\hat{\beta}= \underset{ b=1,...,B }{\mathrm{argmin}} \; BI(\beta_b) \; .
\label{eq:findb}
\end{equation}

\section{Experiments}
 
We use two indices to measure the (dis)similarity between an output image $O$ and the ground truth of the images as performance criteria: i) Rand Index ($RI$), and ii) Adjusted Rand Index ($ARI$) \cite{ARI}, which takes values in $[0,1]$. When the output image $O$ and the ground truth image are identical, the $ARI$ and the $RI$ are equal to $1$. Moreover, the $ARI$ equals $0$ when the $RI$ equals its expected value. 

\begin{table}[h!]
\centering
\caption{Performance of the Algorithms for Thematic Mapper Image}
\begin{tabular}{ccccccccccccc}
& \multicolumn{1}{c}{\textbf{Average Base}} & \multicolumn{1}{c}{\textbf{Algorithm 1}} & \multicolumn{1}{c}{\textbf{DL}} & \multicolumn{1}{c}{\textbf{QD}} \\
$RI$ & 0.703 & 0.704 & 0.710 & 0.714 \\
$ARI$ & 0.159 & 0.160 & 0.184 & 0.174 \\
\end{tabular}%
\label{tab:tmi}%
\end{table}%

In the first set of experiments, we employ the proposed segmentation fusion algorithms on $7$ band Thematic Mapper Image which is provided by MultiSpec \cite{multispec}. We split the image with size $169 \times 169$ into training and test images: i) a subset of the pixels with coordinates $x=(1:169)$ and $y=(1:90)$ is taken as the training image and ii) a subset of the pixels with coordinates $x=(1:169)$ and $y=(91:142)$ is taken as the test image. In the images, there are $C=6$ clusters corresponding to different segments. 

We first implement k-means on $J=7$ different bands, in order to perform multi-modal data fusion. The termination time of Filtered Stochastic BOEM is set to $T=1000$. Assuming that we do not know the number of clusters $C$ in the image, we employ \eqref{eq:findc} using the training data in order to find the optimal $C$ for $c=2,3,4,5,6,7,8,9,10$. Then, we find $\hat{C}=6$ with $ARI=0.2648$. We employ \eqref{eq:findb} for $\Xi= \{ 0.1, 0.2, 0.3, 0.4, 0.5, 0.6, 0.7, 0.8, 0.9, 0.99 \}$ and find $\hat{\beta}=0.9$ with $ARI=0.2648$. The results of the experiments on the test data of Thematic Mapper Image are given in Table \ref{tab:tmi}. In the \textbf{Average Base} column, the average performance values of k-means algorithms are given. The performance values of the segmentation fusion algorithm are given in the column labeled  \textbf{Algorithm 1}. We observe that the performance values of \textbf{Algorithm 1} are similar to the arithmetic average of the performance values of k-means algorithms. The performance of Distance Learning and Quasi-distance Learning algorithms, are given in \textbf{DL} and \textbf{QD}, respectively. Since distance functions for \textbf{Algorithm 1} are computed using the segmentation-wise $ARI$ values in \textbf{DL} and \textbf{QD}, we observe that performance increases in the $ARI$ values of \textbf{DL} and \textbf{QD} compared to \textbf{Algorithm 1}.

In the second set of the experiments, we employ k-means, Graph Cut and Mean Shift algorithms on $7$-band training and test images. Now, the image segmentation problem is considered as a pixel clustering problem in $7$ dimensional spaces. We find $\hat{C}=6$ and $\hat{\beta}=0.9$ with $ARI=0.267$ using the training data. The results on the test data are given in Table \ref{tab:7band}. The performance values of \textbf{Algorithm 1} are closer to the performance values of the \textbf{Mean Shift} algorithm, since the output image of \textbf{Algorithm 1} is closer to the output segmentation of the \textbf{Mean Shift} algorithm. We observe that the $ARI$ values of \textbf{DL} are greater than the values of \textbf{QD}, since \textbf{DL} computes the distance functions by computing the $ARI$ values between the segmentations. However, the $RI$ values of \textbf{QD} are greater than the values of \textbf{DL}, since \textbf{QD} calibrates  distance functions considering the distance measure of the $RI$.
\begin{table}[h!]
\centering
\caption{Experiments using k-means, Graph Cut and Mean Shift Algorithms on $7$-band Images }
\begin{tabular}{ccccccccccccc}
& \multicolumn{1}{c}{\textbf{k-means}} & \multicolumn{1}{c}{\textbf{Graph Cut}} & \multicolumn{1}{c}{\textbf{MeanShift}}  \\
\textit{RI} & 0.715 & 0.717 & 0.714  \\
\textit{ARI} & 0.125 & 0.132 & 0.176  \\
& \multicolumn{1}{c}{\textbf{Algorithm 1}} & \multicolumn{1}{c}{\textbf{DL}} & \multicolumn{1}{c}{\textbf{QD }} \\
$RI$ & 0.714 & 0.710 & 0.724 \\
$ARI$ & 0.176 & 0.180 & 0.178 \\
\end{tabular}%
\label{tab:7band}%
\end{table}%

In the third set of experiments, we employ k-means algorithm on each band of $12$-band Moderate Dimension Image: June 1966 aircraft scanner Flightline C1 (Portion of Southern Tippecanoe County, Indiana) \cite{multispec}. The size of the image is $ 949 \times 220$, and there are $11$ clusters in the ground truth of the image \cite{multispec}. We randomly select $104390$ pixels for training and the remaining $104390$ pixels for testing. We find $\hat{C}=11$ and $\hat{\beta}=0.9$ with $ARI=0.004$ using the training data. The results on the test data are given in Table \ref{tab:mod1} and Table \ref{tab:mod2}. We observe that the performance values for \textbf{Algorithm 1} are smaller than the average performance values of base segmentation outputs. Since the distance functions are computed for each segmentation pair, we achieve better performance for distance learning algorithms (\textbf{DL} and \textbf{QD}). 

\begin{table}[h!]
\centering
\caption{Performance of k-means Algorithms for Moderate Dimension Image}
\begin{tabular}{rrrrrrrrrrrrr}
\multicolumn{1}{c}{} & \multicolumn{1}{c}{\textbf{Ch1}} & \multicolumn{1}{c}{\textbf{Ch2}} & \multicolumn{1}{c}{\textbf{Ch3}} & \multicolumn{1}{c}{\textbf{Ch4}} & \multicolumn{1}{c}{\textbf{Ch5}} & \multicolumn{1}{c}{\textbf{Ch6}} \\
\multicolumn{1}{c}{$RI$} & \multicolumn{1}{c}{0.537} & \multicolumn{1}{c}{0.531} & \multicolumn{1}{c}{0.528} & \multicolumn{1}{c}{0.532} & \multicolumn{1}{c}{0.532} & \multicolumn{1}{c}{0.523}  \\
\multicolumn{1}{c}{$ARI$} & \multicolumn{1}{c}{0.014} & \multicolumn{1}{c}{0.006} & \multicolumn{1}{c}{0.009} & \multicolumn{1}{c}{0.009} & \multicolumn{1}{c}{0.006} & \multicolumn{1}{c}{-0.003}  \\
\multicolumn{1}{c}{} & \multicolumn{1}{c}{\textbf{Ch7}} & \multicolumn{1}{c}{\textbf{Ch8}} & \multicolumn{1}{c}{\textbf{Ch9}} & \multicolumn{1}{c}{\textbf{Ch10}} & \multicolumn{1}{c}{\textbf{Ch11}} & \multicolumn{1}{c}{\textbf{Ch12}} \\
\multicolumn{1}{c}{$RI$} & \multicolumn{1}{c}{0.529} & \multicolumn{1}{c}{0.531} & \multicolumn{1}{c}{0.534} & \multicolumn{1}{c}{0.527} & \multicolumn{1}{c}{0.540} & \multicolumn{1}{c}{0.540} \\
\multicolumn{1}{c}{$ARI$} & \multicolumn{1}{c}{0.000} & \multicolumn{1}{c}{0.008} & \multicolumn{1}{c}{0.015} & \multicolumn{1}{c}{-0.003} & \multicolumn{1}{c}{0.023} & \multicolumn{1}{c}{0.018} 
\end{tabular}%
\label{tab:mod1}%
\end{table}%

\begin{table}[h!]
\centering
\caption{Performance of the Algorithms for Moderate Dimension Image}
\begin{tabular}{ccccccccccccc}
&\multicolumn{1}{c}{\textbf{Average Base}} & \multicolumn{1}{c}{\textbf{Algorithm 1}} & \multicolumn{1}{c}{\textbf{DL}} & \multicolumn{1}{c}{\textbf{QD}} \\
$RI$  & 0.532 &  0.530 & 0.533& 0.530 \\
$ARI$ & 0.009 & 0.007 & 0.011 & 0.011 \\
\end{tabular}%
\label{tab:mod2}%
\end{table}%

\section{Conclusion}

In this study, we introduce a new approach for the fusion of
the segmentation outputs of several segmentation algorithms to achieve a consensus segmentation. Therefore, the output segmentation fusion algorithm
can be interpreted as the image representing the mutual information on a set of segmentation outputs obtained from various segmentation algorithms.
 
We construct the candidate segmentation set by using the k-means, Mean Shift and Graph Cuts methods applied on the hyper-spectral images. The parameter optimization of the segmentation is embedded into the Filtered stochastic BOEM method. Additionally, the distance metrics are learned using the training data in order to enhance the segmentation performance without preselecting parameters, or evaluating the outputs for specific targets. The performances of the suggested segmentation fusion algorithm  demonstrates its efficacy 
in compromising over-segmented results and under-segmented  
results.

\bibliographystyle{IEEEbib}
\bibliography{IEEEabrv,icip_r}

\end{document}